\documentclass{article}

\usepackage{arxiv}

\usepackage[utf8]{inputenc} 
\usepackage[T1]{fontenc}    
\usepackage{hyperref}       
\usepackage{url}            
\usepackage{booktabs}       
\usepackage{amsfonts}       
\usepackage{nicefrac}       
\usepackage{microtype}      
\usepackage{lipsum}         
\usepackage{graphicx}
\usepackage{natbib}
\usepackage{doi}
\usepackage{tikz}  
\usetikzlibrary{arrows, positioning}  
\usepackage{subcaption}
\usepackage{multirow}
\usepackage{amsmath}
\usepackage{cleveref}       
\usepackage{algorithm}
\usepackage{algorithmic}
\usepackage{adjustbox}
\usepackage{amsmath}
\usepackage{amssymb}
\usepackage{multirow}
\usepackage{cleveref}
\usepackage{tabularx}
\usepackage{booktabs} 
\usepackage{todonotes}
\usepackage{array}
\usepackage{tikz}
\usepackage{tcolorbox}
\usepackage{fancyvrb}
\tcbuselibrary{skins, breakable, listingsutf8}
\usetikzlibrary{positioning, shapes.geometric, arrows.meta}
\usepackage{xcolor}
\usepackage{booktabs}
\usepackage{multirow}
\usepackage{amsmath}
\usepackage{booktabs}
\usepackage{pgfplots}
\usepackage{amsmath}
\pgfplotsset{compat=1.17}
\usepackage{listings} 
\usepackage{xcolor}
\usepackage{pgfplots}
\usepackage{pgfplotstable}
\usepgfplotslibrary{groupplots}
\usepackage{tikz}
\usepackage{subcaption}

\title{\LARGE 
FairTabGen: High-Fidelity and Fair Synthetic Health Data Generation from Limited Samples
}
\author{
  Nitish Nagesh\thanks{Equal contribution} \\
  University of California, Irvine \\
  \texttt{nnagesh1@uci.edu}
  \And
  Salar Shakibhamedan\footnotemark[1] \\
  TU Wien (Vienna University of Technology) \\
  \texttt{salar.shakibhamedan@tuwien.ac.at}
  \And
  Mahdi Bagheri \\
  University of California, Irvine \\
  \texttt{mahdib1@hs.uci.edu}
  \And
  Ziyu Wang \\
  University of California, Irvine \\
  \texttt{ziyuw31@uci.edu}
  \And
  Nima TaheriNejad \\
  Heidelberg University \\
  \texttt{nima@uni-heidelberg.de}
  \And
  Axel Jantsch \\
  TU Wien (Vienna University of Technology) \\
  \texttt{axel.jantsch@tuwien.ac.at}
  \And
  Amir M. Rahmani \\
  University of California, Irvine \\
  \texttt{amirr1@uci.edu}
}

\date{}

\renewcommand{\shorttitle}{\textit{arXiv} Template}

\hypersetup{
    pdftitle={FairTabGen: High-Fidelity and Fair Synthetic Health Data Generation from Limited Samples},
    pdfauthor={Nitish Nagesh, Salar Shakibhamedan, Mahdi Bagheri, Ziyu Wang, Nima TaheriNejad, Axel Jantsch, Amir M. Rahmani}
}

\begin{document}
\maketitle

\begin{abstract}
\textcolor{black}{Synthetic healthcare data generation offers a promising solution to research limitations in clinical settings caused by privacy and regulatory constraints. However, current synthetic data generation approaches require specialized knowledge about training generative models and require high computational resources. In this paper, we propose FairTabGen, an LLM-based tabular data generation framework that produces high-quality synthetic healthcare data using only a small subset of the original dataset. Our method combines in-context learning, prompt curation and embedding structural constraints for data synthesis. We evaluate performance on MIMIC-IV dataset. Our method using 99\% less data and achieving 50\% improvement for fairness through unawareness while maintaining competitive predictive utility. However, we observe data distribution of racial groups is skewed affecting  demographic parity. We thereafter apply bias mitigation algorithms in the pre-processing stage, improving overall fairness by 10\% highlighting effectiveness of our approach.
}
\end{abstract}
\begin{keywords}{synthetic data, tabular health data, counterfactual fairness, bias mitigation}
\end{keywords}

\section{Introduction}

\textcolor{black}{Real-world health datasets are prone to systemic biases through protected attributes such as gender, age and race\cite{chen2021ethical}. These biases arise from a multitude of factors including limited health access for certain demographic groups and use of biased clinical proxies\cite{obermeyer2019dissecting}. While current evaluation relies on evaluating predictive performance of machine learning models, high accuracy alone is not sufficient since it can negative impacts on vulnerable groups\cite{ghassemi2020review}. Therefore, there is a need to assess data fairness to ensure equitable health outcomes\cite{rajkomar2018ensuring}.} 

\subsection{Synthetic data techniques}
\textcolor{black}{Advances in generative models such as generative adversarial networks (GANs) and variational autoencoders (VAEs)\cite{qian2023synthcity} and diffusion models\cite{kotelnikov2023tabddpm} have enabled high-fidelity data generation across multiple domains. Further, large language models (LLMs) are being used as data generators in tabular settings\cite{borisov2022language,solatorio2023realtabformer} enabling data augmentation at a larger scale. Synthetic data offers a mechanism to mitigate bias by providing a mechanism to augment data in low resource settings\cite{jordon2022synthetic}.}

\subsection{Bias mitigation via synthetic data}
\textcolor{black}{Synthetic data serves as a mechanism to mitigate bias by imposing fairness constraints in the data generation process\cite{pmlr-v238-abroshan24a}. This in addition to algorithmic approaches for in-, pre- and post- processing techniques\cite{mehrabi2021survey}}.     
\subsection{Research Gap and Research Question}
\textcolor{black}{Current efforts for synthetic data generation\cite{van2021decaf,seedat2023curated} rely on extensive data curation and training generative models which are time consuming and require high computational resources. The challenge therefore is to generate synthetic data that captures underlying data characteristics without further amplifying biases in resource constrained settings.}     
\textcolor{black}{We seek to answer the following question -- \textbf{\textit{"How can we generate synthetic data that retains predictive utility while being fair across sensitive attributes using limited data samples.}}}
\subsection{Contributions}
\textcolor{black}{In this work, we propose \textbf{FairTabGen}, a prompt-based framework that leverages Large Language Models (LLMs) to synthesize fair, high-utility tabular health data under resource constraints. Our contributions are as follows:
    \begin{enumerate}
        \item We present a method to generate tabular synthetic data using curated prompts that incorporate utility and fairness constraints without extensive fine tuning.
        \item We evaluate this approach across multiple metrics including counterfactual fairness, predictive utility and bias mitigation metrics.
        \item We demonstrate the effectiveness of our method on the MIMIC-IV dataset, showing comparable fairness using 99\% less source data while maintaining competitive predictive utility.
    \end{enumerate}
    By evaluating fairness from multiple perspectives and analyzing the impact of different evaluation and testing methods, we aim to provide a more rigorous framework for the responsible use of synthetic data in healthcare.
}

\begin{figure}
    \centering
    \includegraphics[width=0.65\linewidth]{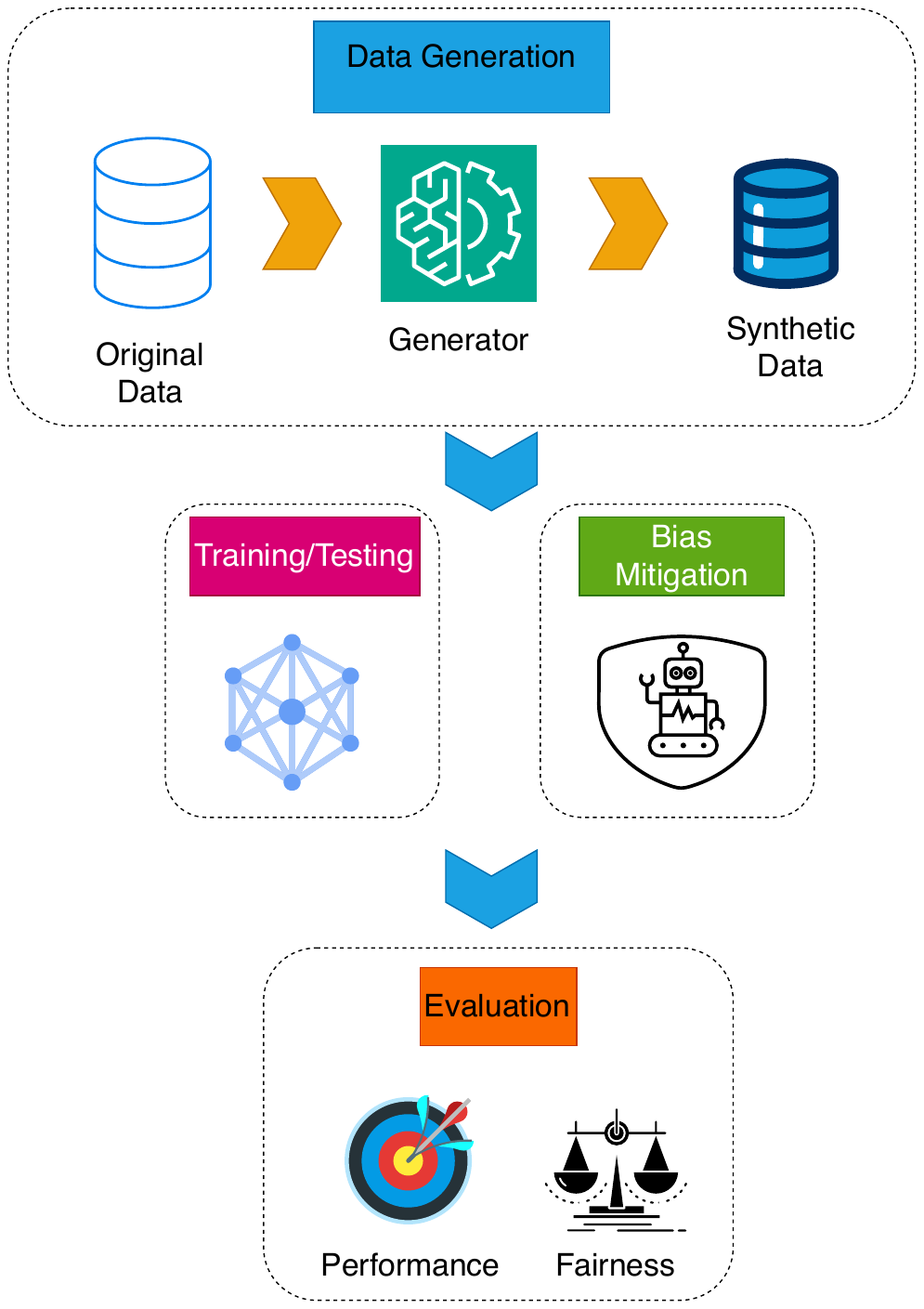}
    \caption{\textcolor{black}{Proposed synthetic data generation architecture. Data is first curated in the data processor, then supplied to the generator which gives the desired output. The evaluator evaluates performance and fairness. Finally, the bias mitigator module runs multiple techniques which are evaluated based on closeness to original value.}}
    \label{fig:high-level-block}
\end{figure}

\section{Preliminaries}
In this section, we define the objective, establish the theoretical foundations for our work, define the core principles of counterfactual fairness, explore bias mitigation algorithms and outline metrics used to quantify algorithmic bias.

\subsection{Objective}
The goal is to generate synthetic data that is matches the distribution to the original data while satisfying fairness constraints for protected attributes and target outcome.

\subsection{Utility Metrics}
We use area under the {receiver operating curve (AUROC)} to measure utility, which is a combination of precision and recall\cite{sajjadi2018assessing}.
{Precision} quantifies how realistic the synthetic data is by calculating the fraction of synthetic points that reside within the manifold of the real data. 
{Recall} measures the diversity of the generated data by calculating the fraction of real data points that fall within the manifold of the synthetic distribution. A higher value of AUROC indicates that the synthetic data and real have have similar distributions.
\subsection{Counterfactual Fairness}
We compute counterfactual fairness\cite{kusner2017counterfactual}
through {Fairness Through Unawareness (FTU)} and {Demographic Parity (DP)}. {FTU} asserts that a model is fair if its prediction remains constant when the protected attribute is flipped, defined as $FTU = \mathbb{E} [ | \hat{y}(x, a) - \hat{y}(x, a') | ]$. Ideally, this value approaches zero. {DP} measures the independence of the prediction and the protected attribute, calculated as the absolute difference in positive prediction rates between groups:
\begin{equation*}
DP = | P(\hat{Y}=1 | A=0) - P(\hat{Y}=1 | A=1) |
\end{equation*}

\subsection{Bias Mitigation Algorithms}
To address systemic disparities, we apply four pre-processing mitigation strategies designed to debias the training data prior to model downstream application:
\begin{itemize}
    \item {Suppression (SUP):} Achieves fairness through unawareness by explicitly excluding sensitive attributes from the feature set \cite{kamiran2012data}.
    \item {Correlation Remover (CoR):} Applies a linear transformation to remove both sensitive attributes and their correlations with non-sensitive features \cite{weerts2023fairlearn}.
    \item {Disparate Impact Remover (DIR):} Edits feature values to ensure distributions are similar across protected and unprotected groups while preserving rank-order \cite{feldman2015certifying}.
    \item {Reweighing (RW):} Assigns weights to training instances based on group membership and target labels to neutralize dependencies between variables \cite{kamiran2012data}.
\end{itemize}

\subsection{Composite Fairness}
To provide a comprehensive assessment of fairness, we utilize metrics that evaluate both the predictive quality and the disparity between groups:
\begin{itemize}
    \item {ABROCA (Area Between ROC Curves):} Quantifies the divergence between the ROC curves of different demographic groups. A value closer to zero indicates higher fairness \cite{gardner2019evaluating}.
    \item {True Positive Rate Difference (TPRD):} Measures the gap in the ratio of correctly identified positive cases between the protected and unprotected groups\cite{hardt2016equality}.
    \item {Error Rate Difference (ERD):} Captures the disparity in the overall misclassification rates between groups, providing insight into whether one group bears a disproportionate burden of model error \cite{berk2021fairness}.
\end{itemize}

We normalize the above metrics to calculate a composite fairness score\cite{liu2025can} 
{\small
\begin{equation*}
    \text{Composite Fairness} = \frac{3 - |ABROCA| - |ERD| - |TPRD|}{3}
\end{equation*}
}

\section{Methodology}
\textcolor{black}{Our methodology implements a modular and iterative framework for high-fidelity and fairness-aware tabular data synthesis for clinical environments.}

\subsection{Proposed Architecture}
\textcolor{black}{The system architecture consists of three stages as shown in Figure~\ref{fig:high-level-block}. The first stage is data curation and pre-processing to extract high quality data samples. The next stage is synthetic data generation, the third stage is multi-dimensional utility and fairness evaluation and the final stage is iterative bias mitigation.}

\subsection{Dataset and Preprocessing}
\textcolor{black}{We use the MIMIC-IV dataset\cite{johnson2021mimic} due to availability of a wide range of clinical features across a diverse population group. Out of the cohort of 418,640 records, we extract 200 records of seed data as in-context samples and perform missing data imputation.}

\subsection{Fairness-aware Synthetic Data Generation}
\textcolor{black}{We leverage an existing prompt structure\cite{seedat2023curated} to integrate curate data samples and supply it to a large language model for data synthesis.}

\subsection{Prompt Structure and Attributes}
\textcolor{black}{The generator is initialized with a structured prompt shown in the table. The roles of clinical variables are defined based on the standard fairness model\cite{plecko2022causal} and is outlined below:
\begin{itemize}
\item \textbf{Sensitive Attribute (
X):} \texttt{race} (binarized to White and Black/African American).
\item \textbf{Confounders (Z):} \texttt{age} and \texttt{gender}, which influence both 
 and the outcome.
\item \textbf{Mediators (W):} 23 clinical features including the Charlson Comorbidity Index, elective status, and 20 frequent diagnostic codes (e.g., ICD codes for heart failure, hypertension, and diabetes).
\item \textbf{Target Outcome (Y):} Length of Stay (\texttt{los\_seconds}), binarized at a 4-day threshold (345600 seconds).
\end{itemize}
The \texttt{header} represents list of all the features and 200 \texttt{in-context samples} are considered. The prompts are processed in batches of 100 samples to accommodate context window limitations and optimize token throughput. To ensure statistical reliability, all experiments are repeated across five independent runs.}

\subsection{Data Synthetis and Training}
\textcolor{black}{We adopt the Train Synthetic, Test Real (TSTR) paradigm\cite{subah2024mitigating,liu2025can}. By training predictive models on generated data and validating them against a held-out test set from the original sample, we can assess robustness of data quality while satisfying fairness constraints.}

\subsection{Bias Mitigation and Evaluation}
\textcolor{black}{We evaluate the synthesized data across three fronts. First, we assess counterfactual fairness through FTU and DP enabling us to understand effects of incorporating fairness constraints during data generation. Next, we evaluate predictive performance through precision, recall and AUROC across different machine learning classifiers. Finally, to address residual bias, we apply pre-processing mitigation techniques (CoR, SUP, DIR, and RW) before model training. The final outcomes are evaluated using ABROCA, TPRD, and ERD to quantify the success of the fairness-utility trade-off in the TSTR context.
}

\subsection{Experimental Setup}
\textcolor{black}{We implement our framework in Python and PyTorch and orchestrate synthetic data using the OpenAI\cite{hurst2024gpt} API using \texttt{gpt-4o} model. We set the temperature to 0.90 and top-p sampling to 0.95. We generate 100 samples per batch to account for context window limitations. We perform intial experiments and data curation on an Intel Core i7-10700K CPU with 16 GB. For training classifiers and data generation at scale, we use the NVIDIA RTX 3090 and 4090 GPUs hosted on an Ubuntu-based environment. We reproduce baselines\cite{seedat2023curated,van2021decaf} to ensure comparative analysis.}

\tcbset{
  mypromptstyle/.style={
    title= Prompt Used for Synthetic Data Generation,
    colback=gray!5!white,
    colframe=black,
    fonttitle=\bfseries,
    sharp corners=southwest,
    enhanced,
    breakable,
    listing only,
    listing options={
      basicstyle=\scriptsize\ttfamily,
      keywordstyle=\color{blue},
      escapeinside=||,
    }
  }
}
\begin{adjustbox}{width=.9\columnwidth}
\begin{tcolorbox}[mypromptstyle,width=0.95\columnwidth]
prompt =\\
System role: You are a tabular synthetic data 
generation model. You are a synthetic data generator.
Your goal is to produce data which mirrors the given
examples in causal fairness within a structural 
causal model (SCM) framework and feature and label 
distributions but also produce as diverse samples as 
possible. I will give you real examples first.

Context: Leverage your knowledge about 
healthcare and causal fairness to generate 
realistic but diverse samples. Generated data should 
consider |{\textcolor{blue}{\{Sensitive Features\}}}| as the sensitive 
attribute (X), |{\textcolor{blue}{\{Confounders\}}}| as the confounders 
(Z), |{\textcolor{blue}{\{Dataset Label\}}}| as the target variable/Outcome 
(Y), and the rest of the features as the mediator 
attribute (W). 

Generated data must be structured to 
allow evaluation of fairness through causal pathways,
capturing both direct and indirect effects of the 
sensitive attribute on the target variable, as well 
as possible confounding influences.

The output should be a markdown code snippet formatted 
in the following schema:
|{\textcolor{blue}{\{Header\}}}|

example data:
|{\textcolor{blue}{\{In-context Samples\}}}|

DO NOT COPY THE EXAMPLES but generate realistic but 
new AND diverse samples which have the correct label 
conditioned on the features.
\end{tcolorbox}
\end{adjustbox}

\begin{table}[b]
    \centering
    \begin{tabular}{lcccc}
         \textbf{Data} & \textbf{White} & \textbf{Black} & \textbf{Female} & \textbf{Male}  \\
         \toprule
         Real & 81.77\% & 18.23\% & 51.67\% & 48.33\% \\
         FairTabGen (Ours) & 51.20\% & 48.98\% & 50.38\% & 49.62\% \\
         \bottomrule
    \end{tabular}
    \caption{\textcolor{black}{Demographic representation of sensitive attributes}}
    \label{tab:placeholder}
\end{table}

\begin{table*}[t]
\centering
\caption{Data quality and counterfactual fairness evaluation across real and synthetic datasets. $\uparrow$ indicates higher is better, $\downarrow$ indicates lower is better.}
\label{tab:combined_quality_fairness}
\resizebox{\columnwidth}{!}{%
\begin{tabular}{llccc cc}
\toprule
\textbf{Dataset} & \textbf{Method} 
& \multicolumn{3}{c}{\textbf{Data Quality} $\uparrow$} 
& \multicolumn{2}{c}{\textbf{Counterfactual Fairness} $\downarrow$} \\
\cmidrule(lr){3-5} \cmidrule(lr){6-7}
& & Prec. & Recall & AUROC 
& FTU & DP \\
\midrule
\multirow{4}{*}{MIMIC-IV}
& Original       & $0.701 \pm 0.006$ & $0.647 \pm 0.006$ & $0.851 \pm 0.003$ & $0.042 \pm 0.003$ & $0.065 \pm 0.013$ \\
& DECAF          & $0.670\pm001$ & $0.600\pm002$ & \boldmath{$0.816\pm.001$} & $0.097\pm0.001$ & $0.012\pm0.001$ \\
& CLLM           & $0.684 \pm 0.005$ & $0.736 \pm 0.002$ & $0.639 \pm 0.001$ & $0.081 \pm 0.002$ & $0.024 \pm 0.001$ \\
& {FairTabGen} & \boldmath{$0.905 \pm 0.003$} & \boldmath{$0.969 \pm 0.008$} & $0.644 \pm 0.003$ & \boldmath{$0.025 \pm 0.003$} & \boldmath{$0.001 \pm 0.005$} \\
\bottomrule
\end{tabular}
}
\end{table*}

\section{Results and Discussion}

\textcolor{black}{In this section, we evaluate the synthetic data distribution, it's predictive utility and effectiveness across fairness metrics for the MIMIC-IV dataset.}
\subsection{Data Distribution and Representation}
\textcolor{black}{The original data is skewed towards the majority racial group while gender is relatively balanced. 
In our synthesis, the generated data for the majority racial group decreased by 30.27\%, while the representation for the minority group increased by 30.75\%. In contrast, gender distribution remained highly consistent with less than 1\% variation between male and female groups. While racial representation in generated samples is balanced across groups under consideration, the ensuing results should be interpreted within this context.
}

\subsection{Model Selection and Baseline Evaluation}
\textcolor{black}{We evaluate original data across five classifiers Support Vector Machines (SVM), Random Forest (RF), Logistic Regression (LR), Decision Trees (DT), and XGBoost (XGB) to ascertain robustness of prediction performance and fairness reliability. As shown in Table~\ref{tab:baseline_metrics}, XGBoost had best performance utility. For FTU and DP metrics, ensemble methods exhibit comparable values unlike logistic regression and random forest, due to their ability to handle high dimensional features.
Therefore, we use XGBoost as the primary classifier for training and evaluating data generation frameworks.}

\begin{table}[htbp]
\centering
\small
\setlength{\tabcolsep}{4pt} 
\caption{Baseline Utility and Fairness on Real MIMIC Data.}
\label{tab:baseline_metrics}
\begin{tabular}{l ccc c cc}
\toprule
\multirow{2}{*}{\textbf{Model}} & \multicolumn{3}{c}{\textbf{Utility} $\uparrow$} & & \multicolumn{2}{c}{\textbf{Fairness} $\downarrow$} \\
\cmidrule(lr){2-4} \cmidrule(lr){6-7}
& Prec. & Rec. & AUC & & FTU & DP \\
\midrule
DT  & 0.61 & 0.62 & 0.69 & & 0.06 & 0.05 \\
LR  & 0.69 & 0.50 & 0.78 & & 0.00 & 0.02 \\
RF  & 0.71 & 0.64 & 0.85 & & 0.07 & 0.08 \\
SVM & 0.76 & 0.25 & 0.64 & & 0.00 & 0.05 \\
XGB & 0.70 & 0.65 & 0.86 & & 0.04 & 0.07 \\
\bottomrule
\end{tabular}
\end{table}

\subsection{Utility and Counterfactual Fairness Analysis}
\textcolor{black}{We evaluate utility and counterfactual fairness of our approach against existing GAN-based\cite{van2021decaf} and LLM-based\cite{seedat2023curated} data generation techniques using the the XGBoost classifier under the train synthetic test real (TSTR) paradigm. Relative to original data, our method exhibits superior precision and recall while DECAF has higher AUROC as shown in Table~\ref{tab:combined_quality_fairness}. Though the overall probability distribution is better captured by DECAF, our approach enables more accurate representation of positive clinical outcomes.}
\textcolor{black}{Moreover, our approach achieves lower FTU and DP compared to existing generators enabling usage for training downstream clinical models.}

\subsection{Bias Mitigation and Composite Fairness}
\textcolor{black}{Ultimately, we evaluate effectiveness of pre-processing mitigation algorithms such as SUP, COR, DIR and RW on individual and composite fairness metrics. Figure~\ref{fig:bias-mitigation-plots} illustrates the directional deviation ($\Delta$) of synthetic data from the real-world baseline. For RW, COR, our method shows very limited deviation for ERD and TPRD. For composite fairness, our method has the smallest variation across all bias mitigation algorithms compared to original data showcasing the validity of our proposed framework.
}

\pgfplotsset{compat=1.18}
\begin{figure*}[t] 
    \centering
    \begin{tikzpicture}
       \begin{groupplot}[
            group style={
                group size=2 by 2, 
                vertical sep=1.5cm, 
                horizontal sep=1.8cm
            },
            ybar, 
            /pgf/bar width=6pt, 
            height=5.5cm, width=0.45\textwidth,
            symbolic x coords={None, SUP, COR, DIR, RW},
            xtick=data,
            ymajorgrids=true, 
            grid style={dashed, gray!30},
            ymin=-0.2, ymax=0.25,
            ylabel style={font=\small},
            title style={font=\small\bfseries},
            xticklabel style={font=\footnotesize},
            yticklabel style={font=\footnotesize},
            legend style={at={(0.5,-0.35)}, anchor=north, legend columns=-1, font=\small}
        ]

        \nextgroupplot[title={(a) ABROCA Deviation ($\Delta$)}, ylabel={$\Delta$ ABROCA}]
        \addplot[fill=red!70!black] coordinates {(None,0.061) (SUP,0.072) (COR,0.053) (DIR,0.001) (RW,0.092)};
        \addplot[fill=blue!70!black] coordinates {(None,-0.015) (SUP,0.020) (COR,0.020) (DIR,-0.010) (RW,-0.017)};
        \addplot[fill=green!60!black] coordinates {(None,0.007) (SUP,0.033) (COR,-0.020) (DIR,0.045) (RW,0.003)};

        \nextgroupplot[title={(b) ERD Deviation ($\Delta$)}, ylabel={$\Delta$ ERD}]
        \addplot[fill=red!70!black] coordinates {(None,-0.093) (SUP,-0.099) (COR,0.034) (DIR,-0.088) (RW,-0.118)};
        \addplot[fill=blue!70!black] coordinates {(None,-0.041) (SUP,-0.076) (COR,-0.050) (DIR,-0.050) (RW,-0.029)};
        \addplot[fill=green!60!black] coordinates {(None,-0.030) (SUP,-0.030) (COR,-0.005) (DIR,-0.039) (RW,-0.005)};

        \nextgroupplot[title={(c) TPRD Deviation ($\Delta$)}, ylabel={$\Delta$ TPRD}]
        \addplot[fill=red!70!black] coordinates {(None,0.163) (SUP,0.191) (COR,-0.072) (DIR,-0.065) (RW,0.154)};
        \addplot[fill=blue!70!black] coordinates {(None,0.019) (SUP,0.041) (COR,0.008) (DIR,0.022) (RW,0.024)};
        \addplot[fill=green!60!black] coordinates {(None,-0.004) (SUP,-0.014) (COR,-0.013) (DIR,-0.013) (RW,-0.030)};

        \nextgroupplot[
            title={(d) Comp. Fairness Deviation ($\Delta$)}, 
            ylabel={$\Delta$ Fairness},
            legend style={at={(-0.1,-0.35)}, anchor=north, legend columns=-1, font=\small}
        ]
        \addplot[fill=red!70!black] coordinates {(None,-0.081) (SUP,-0.096) (COR,-0.057) (DIR,-0.033) (RW,-0.108)};
        \addplot[fill=blue!70!black] coordinates {(None,0.010) (SUP,-0.021) (COR,-0.010) (DIR,-0.002) (RW,0.001)};
        \addplot[fill=green!60!black] coordinates {(None,0.005) (SUP,-0.006) (COR,-0.025) (DIR,-0.014) (RW,-0.009)};
        \legend{DECAF, CLLM, \textbf{Ours (FairTabGen)}}

        \end{groupplot}
    \end{tikzpicture}
    \caption{Distributional Drift Analysis. Bars represent the signed deviation from the real-world baseline ($\Delta = \text{Method} - \text{Real}$). Subplot (d) illustrates the {Composite Fairness Deviation}, where FairTabGen consistently maintains the highest stability (closest to zero) across all mitigation strategies compared to state-of-the-art baselines.}
    \label{fig:bias-mitigation-plots}
\end{figure*}

\section{Conclusion and Future Work}
We propose a method for generating tabular synthetic data that leverages large language models (LLMs) while meeting fairness constraints. We demonstrate the effectiveness of our approach using limited seed data and structural constraints. By incorporating these constraints within the generation process and using less than 1\% of the source data, we achieved fairness gains while maintaining comparative performance. Furthermore, across multiple bias mitigation algorithms, the composite fairness score is higher, enabling the use of synthetically generated data for training clinical models in the future. 
Despite these promising results, our work has limitations. The data distribution across racial groups is skewed and does not capture the full range of clinical features. Moreover, reliance on black-box models such as GPT-4o limits reproducibility in clinical environments. To address these limitations, we aim to extend our framework to compare across multiple baselines, incorporate open-source models, and generate data at a larger scale to assess predictive utility and fairness. By doing so, we will contribute to an extensible, lightweight framework for synthetic data generation.

\bibliographystyle{unsrt}
\bibliography{ref}

\end{document}